# Biomedical Knowledge Graph: A Survey of Domains, Tasks, and Real-World Applications

Yuxing Lu[1,#], Sin Yee Goi[2,#], Xukai Zhao[3], Jinzhuo Wang[1,*]

*1 Department of Big Data and Biomedical AI, College of Future Technology, Peking University, Beijing, 100871, China*
*2 School of Life Sciences, Peking University, Beijing, 100871, China*
*3 School of Architecture, Tsinghua University, Beijing, 100084, China*

# These authors contributed equally to this work.
* Correspondence should be addressed to: Jinzhuo Wang, email: wangjinzhuo@pku.edu.cn

## Abstract

Biomedical knowledge graphs (BKGs) have emerged as powerful tools for organizing and leveraging the vast and complex data found across the biomedical field. Yet, current reviews of BKGs often limit their scope to specific domains or methods, overlooking the broader landscape and the rapid technological progress reshaping it. In this survey, we address this gap by offering a systematic review of BKGs from three core perspectives: domains, tasks, and applications. We begin by examining how BKGs are constructed from diverse data sources, including molecular interactions, pharmacological datasets, and clinical records. Next, we discuss the essential tasks enabled by BKGs, focusing on knowledge management, retrieval, reasoning, and interpretation. Finally, we highlight real-world applications in precision medicine, drug discovery, and scientific research, illustrating the translational impact of BKGs across multiple sectors. By synthesizing these perspectives into a unified framework, this survey not only clarifies the current state of BKG research but also establishes a foundation for future exploration, enabling both innovative methodological advances and practical implementations.

# Introduction

Knowledge Graphs (KGs) are structured representations of information that capture entities and their relationships within a graph-based data model, thereby enabling seamless integration and analysis of large, heterogeneous datasets. Since Google introduced the concept in 2012, KGs have been applied across various fields—including e-commerce
, scientific research [1], and healthcare [2]. Their capacity for semantic integration and knowledge representation has made them particularly valuable tools for knowledge-intensive fields. Biomedical research generates vast amounts of complex data, spanning molecular interactions, pharmacological insights, and clinical records [3]. Traditional data management strategies, such as relational databases and rule-based systems, struggle to capture the intricate and interdependent nature of biological processes. As a result, valuable insights often remain fragmented, limiting their practical impact on research and healthcare decision-making.

To address these challenges, Biomedical knowledge graphs (BKGs) have emerged as a transformative approach. By integrating diverse data sources, BKGs enable more comprehensive representation and analysis of biomedical knowledge [4, 5]. This structured integration facilitates advanced applications such as disease modeling, drug discovery, and personalized medicine in ways that traditional approaches cannot easily support. Recent advancements in artificial intelligence (AI), particularly in large language models (LLMs) [6] and multi-modal data integration [7], have significantly enhanced the development and utility of BKGs. Techniques such as graph representation learning and natural language processing have expanded the scope and impact of BKG applications, generating novel insights into disease mechanisms and accelerating drug development efforts.

Despite the growing interest in BKGs, existing surveys often focus narrowly on specific domains or methodologies, overlooking the latest technological developments and emerging applications. For example, Nicholson et al. [8] review BKG construction methods, Zeng et al.'s [9] explore BKG applications in drug discovery, and Wu et al.'s [10] as well as Cui et al.'s [11] provide surveys of BKGs in healthcare. This survey aims to fill this gap by providing a holistic review, systematically introducing their development, methodologies, and implications for future biomedical research.

In this paper, we present a systematic survey of BKGs that span three central dimensions: domains, tasks, and applications (**Figure 1**). Our primary contributions are as follows:
1) **Domain-Focused Classification**. We provide a detailed classification of BKGs based on their underlying data sources, ranging from multi-omics to pharmacology and clinical data, highlighting how these domains converge and diverge in practice.
2) **BKG Tasks Illustration**. We examine the key tasks BKGs support, including knowledge management, retrieval, reasoning, and interpretation, emphasizing their foundational role in advancing biomedical research.
3) **Real-World Applications**. We showcase the transformative potential of BKGs in precision medicine, drug discovery, and scientific research, illustrating their tangible impact across biomedical science and patient care.

4) **Emerging Trends and Future Directions**. We analyze current limitations of BKGs, discuss the influence of recent technological breakthroughs (e.g., LLMs and multi-omics integration), and offer insights on how future developments might shape the field.

The remainder of this paper is organized as follows. We begin with an overview of BKG definitions and construction workflows, laying the groundwork for understanding their development and utility. We then discuss the key domains contributing to BKG construction and application, followed by an analysis of core tasks. Next, we illustrate real-world applications, emphasizing the benefits of BKGs in biomedical research and clinical practice. We conclude with a discussion of different data sources, BKGs and tools. Current limitations, and future opportunities, providing a comprehensive outlook on this rapidly evolving field.

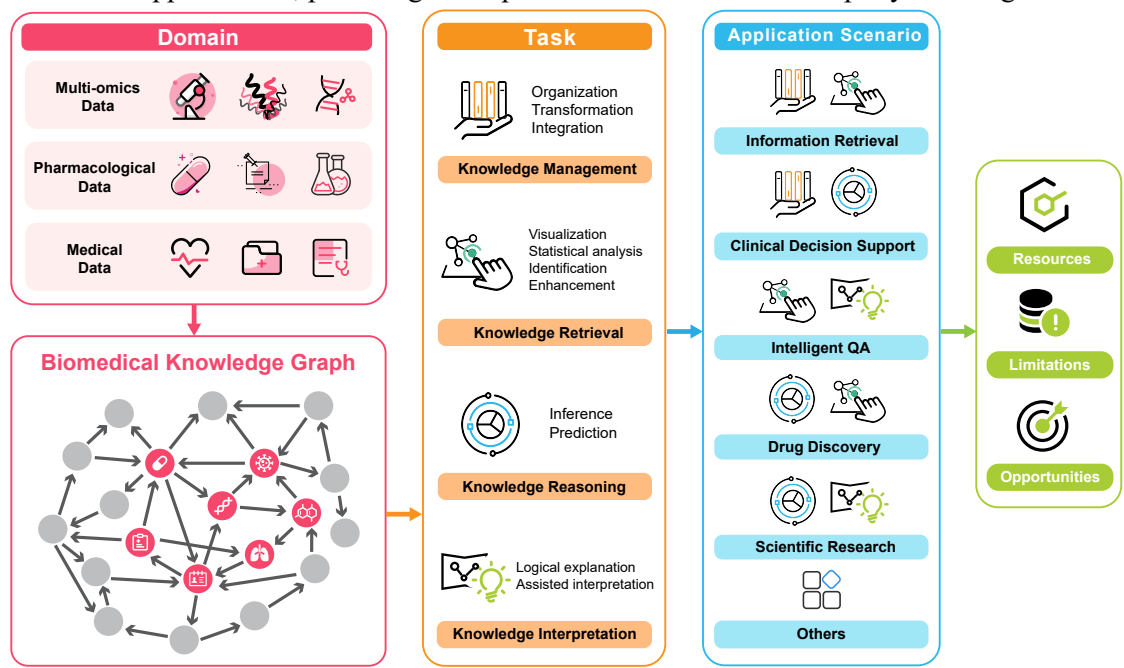

**Figure 1. Overview of BKGs, encompassing their domains, tasks, and applications while addressing the limitations, opportunities, and available resources.**

# Background

## Definition

A BKG is a structured framework designed to represent entities and their relationships within the biomedical domain [12]. Formally, a BKG is defined as $G = (E, R, F)$, where $E$ is the set of entities, $R$ is the set of relations, and $F \subseteq E \times R \times E$ is the set of factual triples. Each triple $(h, r, t) \in F$ consists of a head entity $h \in E$, a relation $r \in R$, and a tail entity $t \in E$, representing facts such as "Gene encodes Protein" or "Drug targets Disease."

BKGs provide a semantic and structural foundation for integrating heterogeneous biomedical data, including genomic information, clinical records, and pharmacological properties. The use of standard vocabularies (e.g., Gene Ontology, SNOMED CT) ensures interoperability, while the graph structure facilitates reasoning and discovery. The graph can also be enriched with weights or probabilities $w: F \to [0, 1]$ to capture the uncertainty or strength of relationships. For example, the plausibility of a triple $(h, r, t)$ can be evaluated using a scoring function $f_r(h, t)$, where higher values indicate greater confidence in the relationship.

## Construction workflow of BKGs

The construction of a BKG is a multi-step process requiring careful design and validation to ensure the graph's quality, consistency, and applicability. The workflow includes data acquisition, preprocessing, relationship extraction, integration, and storage [13-15].

### Data Acquisition and Preprocessing

Biomedical data are collected from a variety of sources, including structured databases $D_s$, semi-structured repositories $D_{ss}$, and unstructured text $D_u$, and other modalities, such as images and omics profiles. These datasets These datasets encompass molecular interactions, disease phenotypes, and drug properties. Preprocessing involves steps like noise reduction, normalization, and deduplication. Mathematically, given raw data $R$, the preprocessed data $D$ is obtained as: $D = \Phi(R)$, where $\Phi$ is a preprocessing pipeline consisting of normalization, mapping, and filtering functions.

### Relationship Extraction and Integration

Relations $R$ are extracted using rule-based systems, machine learning models, or advanced Natural Language Processing (NLP) techniques to identify semantic relationships [16-20]. The extracted relationships are represented as triples $(h, r, t)$, with a scoring function $f_r(h, t)$ quantifying their plausibility. The integration process aligns entities and relations across datasets using ontology matching and schema alignment. The final unified graph $G$ is constructed as:

$$G = \cup_{i=1}^{n}(E_i, R_i, F_i),$$

where $(E_i, R_i, F_i)$ are the components of the $i$-th dataset.

### Graph Validation and Storage

Graph consistency is ensured by enforcing integrity constraints, such as type compatibility for relations. For instance, a relation $r \in R$ may require specific types for the head $h$ and tail $t$ entities, represented as $Type(h) \rightarrow Type(t)$. The completed BKG is stored in graph databases $\mathcal{DB}_G$, with support for querying using languages like SPARQL or Cypher.

### Knowledge Graph Representation Methods

Once constructed, BKGs require efficient representation methods to facilitate reasoning, querying, and downstream applications. Knowledge Graph Embedding (KGE) methods encode entities and relations into continuous vector spaces while preserving graph structure and semantics [4, 13, 21-23].

### Translational Models

Translational models, such as TransE [24], TransH [25], TransD [26], and TransR [27], embed entities and relations into $\mathbb{R}^d$ and enforce translational consistency:

$$\boldsymbol{h} + \boldsymbol{r} \approx \boldsymbol{t},$$

where $\boldsymbol{h}, \boldsymbol{r}, \boldsymbol{t} \in \mathbb{R}^d$ are the embeddings of head, tail, and relation, respectively. The plausibility of a triple is scored as:

$$f_r(h, t) = -||\boldsymbol{h} + \boldsymbol{r} - \boldsymbol{t}||_p,$$

where $||\cdot||_p$ denotes the $p$-norm. Extensions like TransH project entities onto relation-specific hyperplanes, enhancing their ability to model diverse relation types.

### Semantic Matching Models

Models like DistMult [28], and RESCAL [29] decompose the adjacency tensor of the graph, uncovering intricate interaction patterns among entities and their relationships. These models evaluate the compatibility of $(h, r, t)$ based on semantic similarity. A common approach is the bilinear scoring function:

$$f_r(h, t) = \boldsymbol{h}^T \boldsymbol{M}_r \boldsymbol{t},$$

where $\boldsymbol{M}_r \in \mathbb{R}^{d \times d}$ is a relation-specific matrix. Models like DistMult simplify $\boldsymbol{M}_r$ to a diagonal matrix, while RESCAL uses a full rank $\boldsymbol{M}_r$ for greater flexibility.

### Graph Neural Networks

Graph Neural Networks (GNNs), such as Graph Convolutional Networks (GCNs) [30] or Graph Attention Networks (GATs) [31], leverage graph topology to learn context-aware embeddings by aggregating information from neighboring nodes in the graph. The hidden state of entity $i$ at layer $l + 1$ is computed as:

$$\boldsymbol{h}_i^{(l+1)} = \sigma(\sum j \in N_i \frac{1}{c_{i,j}} \boldsymbol{W}^l \boldsymbol{h} j^l + \boldsymbol{b}^l),$$

where $N_i$ denotes the neighbors of $i$, $\boldsymbol{W}^l$ is a learnable weight matrix, and $c_{i,j}$ is a normalization constant, and $\sigma$ is a non-linear activation function. This method captures both local and global graph structure, enabling complex reasoning.

**Embedding in Non-Euclidean Spaces**

Recent advancements in KGE explore embeddings in hyperbolic and complex spaces. For instance, RotatE [32] models relations as rotations in a complex vector space:

$$\boldsymbol{t} = \boldsymbol{h} \circ \boldsymbol{r},$$

where ∘ denotes the Hadamard product. These approaches are particularly effective for capturing hierarchical and cyclic patterns in biomedical data.

**Hybrid Approaches**

In many real-world scenarios, BKGs encompass not only graph structure but also multimodal data like images, omics profiles, or clinical variables [6, 31, 33-35]. Hybrid methods that fuse graph embeddings with these additional data streams can yield richer, more comprehensive representations [10, 36-40]. Heterogeneous network embeddings or meta-path analyses further extend this concept by accounting for diverse entity types and complex multi-step relationships [2, 10, 12, 22, 37, 41-44].

By summarizing the latent structure of the BKG, representation learning techniques enable powerful downstream applications, including link prediction (e.g., suggesting novel drug targets), node classification (e.g., categorizing diseases by etiology), and knowledge inference (e.g., identifying potential molecular pathways) [4, 10, 12, 37, 45-47]. Integrating with large language models and multimodal data holds promise for future innovation, potentially accelerating discovery and clinical translation in personalized medicine [6, 14, 48-51].

# Domains

BKGs serve as comprehensive frameworks that integrate diverse healthcare-related information across various interconnected domains. They provide a structured approach to organize multidisciplinary knowledge in biomedical sciences, encompassing fields such as genomics, drug discovery, and clinical data. By unifying these domains, BKGs support advanced analyses and facilitate interdisciplinary research.

In this section, BKGs are classified into three primary domains: multi-omics, pharmacology, and medical. They can be illustrated as overlapping sets (**Figure 2a**), with non-overlapping regions representing domain-specific tasks and intersecting areas reflecting interdisciplinary collaboration. The intersection of all three domains constitutes a multi-domain BKG, capturing the complex interdependencies within the biomedical field.

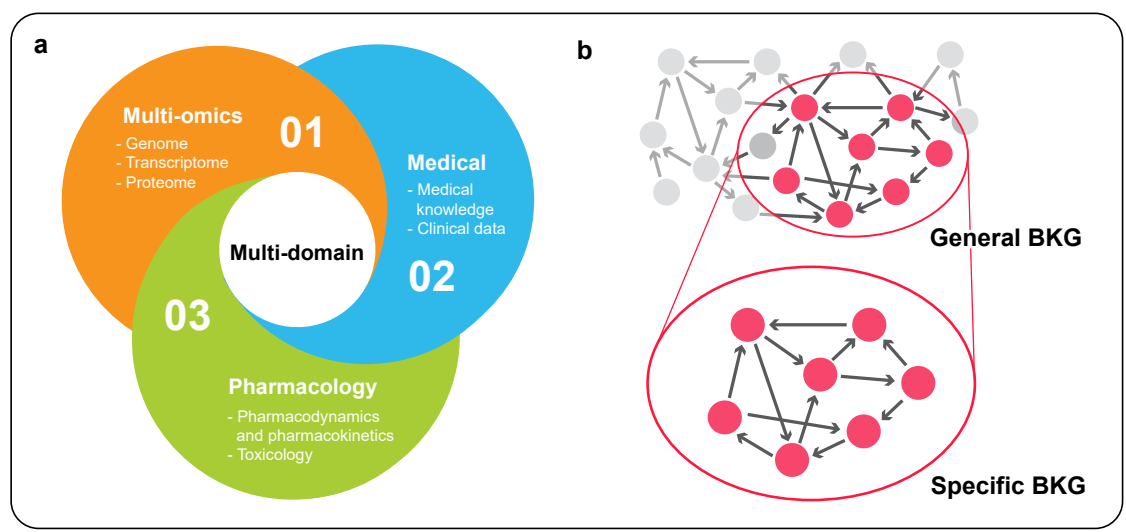

**Figure 2. BKG Domains. a)** The three main domains of BKG, with an overlapping region representing multi-domain BKG; **b)** illustration of general and specific BKG, derived from a complete BKG.

## Multi-omics

Multi-omics BKGs integrate one or more omics data types to enable comprehensive biological analysis and the discovery of novel associations among biological entities. These BKGs focus on three primary areas: genome [52, 53], transcriptome [54-58], and proteome [58-63]. Recent advancements have also led to the development of multi-omics BKGs related to metabolomics [64] and epigenomics [52].

### Genome

Genome-centric BKGs model relationships between genomic entities to investigate disease mechanisms and biological pathways. For example, Feng et al. constructed a genome BKG that enables genomic information retrieval, including positional relationships between genomic regions and specific entities [52]. Jha et al. proposed GenomicsKG, a poly-omics BKG focusing on cancer genomics, which illustrates cancer evolution mechanisms and highlights gene roles and mutation frequencies [53].

### Transcriptome

The transcriptome includes all RNA molecules transcribed from the genome, including both coding and non-coding RNAs. Transcriptomic BKGs tackle the challenges of RNA diversity by integrating RNA-related data. Cavalleri et al. developed RNA-KG which supports the study of RNA interactions and pathophysiological processes such as miRNA-lncRNA associations [54]. Similarly, Liu et al. utilized transcriptomic data to build a BKG for studying Polycystic Ovary Syndrome, identifying pathogenesis similarities across subtypes [55].

### Proteome

Protein-centric BKGs focus on understanding protein structure, function, and interactions. Zhang et al. introduced ProteinKG25, a BKG supporting protein representation pre-training based on Gene Ontology [59]. Binder et al. created an Alzheimer-specific protein BKG to predict disease-protein associations [60]. Similarly, Papadopoulos et al. developed IDP-KG to study intrinsically disordered proteins involved in cell signaling [61]. In addition, Nováček et al. proposed a KG from phosphorylation data to predict kinase-substrate relationships [62].

### Pharmacology

Pharmacology is the study of drugs and medications, focusing on the interactions between drugs and biological system. The pharmacological domain of BKG comprises two main areas: pharmacodynamics and pharmacokinetics [36, 41, 65-74], and toxicology [39, 75-79]. The domain of pharmacodynamics and pharmacokinetics interprets the relationship between drug dosage and its response [80], while the toxicology domain reveals mechanisms of adverse drug effects (ADRs) occurring in the body.

### Pharmacodynamics and pharmacokinetics

Pharmacodynamics examines drug effects on the body, whereas pharmacokinetics studies how the body processes drugs. BKGs in this area model drug-disease associations (DDAs) and drug-target interactions (DTIs) [4, 9]. For instance, Tanaka et al. developed IPM-KG to predict drug pathways [65]; Gonzalez-Cavazos et al. constructed DrugMechDB to promote the understanding between drugs and diseases [66]. Zhang et al. utilized COVID-19-specific BKGs to identify potential treatments through literature-based discovery (LBD) [67]. Similarly, Geleta et al. [68] and Gao et al. [69] constructed drug KGs to discover existing drugs for new diseases by identifying new targets. Additionally, the OREGANO KG proposed by Boudin et al. [70] incorporates natural compounds such as herbs and plant remedies into a BKG aimed for drug repurposing. Therefore, the structured drug information represented in BKG can be considered as an integration of DDA and DTI networks. Aside from that, datasets related to the absorption, distribution, metabolism and excretion of drugs are collected for pharmacokinetics usage. Chen et al. developed a transporter proteins KG to enhance drug discovery process from a pharmacokinetics perspective [71].

### Toxicology

Toxicology-focused BKGs model adverse drug reactions (ADRs) and drug-drug interactions (DDIs) [4, 9]. Abdelaziz et al. proposed a DDI prediction framework, by utilizing a drug BKG where toxicogenomic datasets were integrated into a BKG, generating features such as side effect-based similarity measures [75]. Furthermore, M. Bean et al. developed a BKG to compile adverse reaction datasets for ADR prediction, ensuring patient safety. Talukder et al. mined drug indications and contradictions from public repositories and compiled it into a drug interaction KG [77]; whereas Faruggia et al. integrated drug-related concepts such as drugs, side effects and DDIs into a BKG to distinguish interacting and non-interacting drug pairs [39].

**Medical**

The medical domain integrates healthcare-related information, including prevention, diagnosis, and treatment of diseases. Medical BKGs typically focus on two areas: medical knowledge [33, 81-89] and clinical data [31, 33, 82, 83, 87, 90-93]. Medical knowledge includes medical theories and terminologies, while clinical data consists of real-world data such as patient registries and diagnosis reports. These data can be compiled into medical BKGs as entities and relations, linking patients with similar conditions.

**Medical knowledge**

Medical knowledge encompasses information such as medical concepts, terminologies, and theories that are primarily gathered from medical knowledge databases and scientific publications [10]. Huang et al. developed a medical KG by standardizing medical instructions related to drug responses and disease complications, facilitating the retrieval of medical entities with specific keywords [81]. Shang et al. introduced a clinical KG enriched with domain-specific ontologies, designed for integration with patient information models [82]. Similarly, Gong et al. constructed a KG which linked medical entities with electronic medical record (EMR) datasets, enabling accurate predictions of associations between patients and medications [83]. Moreover, disease-specific KGs further specialize in capturing knowledge about particular illnesses. Zhu et al. created SDKG-11, featuring curated data on 11 diseases from biomedical literature and databases, supporting the discovery of novel insights by visualizing disease-specific patterns and relationships [84]. Li et al. prioritized hepatocellular carcinoma [85], while Huang et al. focused on depression [86], both utilizing extensive datasets to offer deep insights into these diseases.

**Clinical data**

Clinical data consist of real-world patient information critical for diagnostics, treatment planning, and healthcare decision-making. These data are often stored in standardized systems such as EMRs and electronic health records (EHRs). Clinical data in BKGs can be classified into patient registries [31, 82, 83, 87, 90-92], laboratory results [82, 93], and medical images [31, 33].

Patient registries collect comprehensive data about patients, including personal details, symptoms, diagnosis results, and treatment outcomes. Rotmensch et al. extracted disease, symptom and personal attributes such as sex and weight from EMRs to construct a health KG for precise predictions [90]. Zheng et al. proposed a multi-modal KG integrating doctor-patient dialogues to capture a patient's medical history [31]; while Zhen et al. developed a clinical trials KG representing trials, conditions, and outcomes [91]. Laboratory test data provide objective health indicators essential for identifying and monitoring medical conditions. Wang et al.

constructed a diabetes KG which combined test results and lifestyle evidence to identify risk factors and predict health outcomes [93]. Shang et al. extended this concept by developing an EHR-oriented KG where test results are organized as nodes, assisting in disease diagnosis through standardized data integration [82]. Furthermore, Medical imaging data, such as optical coherence tomography (OCT), X-rays, and CT scans, are valuable for enriching BKGs with diagnostic insights. Gao et al. organized ophthalmic medical images and corresponding reports into a KG, creating a centralized system for managing ophthalmology records [33]. This multi-modal KG connects individual medical records via shared attributes such as gender and medical history. Zheng et al. further enhanced this approach by integrating chest X-rays, ultrasounds, and CT scans with doctor-patient dialogues into a COVID-19 KG, linking radiological findings with textual data to improve interpretation [31].

**Multi-domain**

Multi-domain BKGs represent the intersection of the multi-omics, pharmacology, and medical domains, integrating datasets from one or more of these areas. These overlapping regions capture the interconnected nature of biomedical science, providing a comprehensive representation of biological interactions and supporting cross-disciplinary research. By unifying diverse datasets, multi-domain BKGs enable the modeling of complex interrelationships across the biomedical field.

In the omics-medical domain, Santos et al. developed the CKG, which combines clinical and molecular proteomic data to interpret disease phenotypes and identify protein biomarkers [1]. Similarly, Renaux et al. curated genetic and clinical datasets to construct BOCK for oligogenic diseases, integrating disease information and biological networks to predict pathogenic gene interactions [94]. Next, BKGs in the omics-pharmacology domain merge omics and drug datasets to explore alternative treatments. Gao et al. proposed a BKG for drug-drug interaction (DDI) prediction, integrating protein-drug triples from medical documents and leveraging protein-protein interactions (PPIs) to uncover novel DDIs [95]. In the pharmacology-medical domain, BKGs emphasize the integration of drug and disease information to support treatment recommendations. Huang et al. introduced a COVID-19 BKG featuring drug-disease associations (DDAs) and disease-chemical interactions, aiding therapeutic development [96]. Additionally, BKGs that span all three domains - multi-omics, pharmacology, and medical, are considered the most comprehensive BKGs, encompassing the full spectrum of biomedical sciences [97-100]. Notable examples include PharmKG by Zheng et al., which harmonizes gene, drug, and disease information [97]; and TarKG by Zhou et al., which improves access to heterogeneous biomedical datasets [98].

Apart from that, a multi-domain BKG can be further classified into two types [45, 84]: general BKG [97-100] and specific BKG [38, 55, 67, 74, 84-86, 93, 96, 101, 102]. Both types are compiled from multi-domain datasets, but they differ in scope and purpose (**Figure 2b**). General BKGs are constructed from a wide array of datasets, encompassing a broad spectrum of biological data. They are model complex biological interactions, providing a comprehensive view of the biological system. Examples of general BKGs include Hetionet [99] and PharmKG

[97], both of which incorporate entities such as genes, drugs, and diseases, to support various biomedical research applications. Specific BKGs focus on narrower topics, such as particular diseases or biological pathways, using targeted datasets. They are subsets of general BKGs and are typically smaller in scale, enabling detailed exploration of specialized topics. For instance, Zhu et al. developed a BKG for rare diseases, facilitating the discovery of orphan drugs [101]. Huang et al. created a Kawasaki disease BKG [102], and Nian et al. proposed a BKG for Alzheimer's disease [74]. Disease-specific BKGs have also been applied to COVID-19 research, aiding treatment and drug development efforts [38, 67, 96].

While general BKGs provide a versatile foundation for broad biomedical research, specific BKGs offer focused insights into particular areas of interest, making them valuable for both hypothesis generation and targeted investigations.

# Tasks

BKG enables diverse biomedical applications through multiple essential tasks that serve as the functional components of its application system. Capable in handling typical database functions, BKG's graph topology allows it to go beyond conventional databases. In this section, we classify BKG tasks into four main categories (**Figure 3**): knowledge management, knowledge retrieval, knowledge reasoning, and knowledge interpretation. Knowledge management aligns the disparate data from various sources, enhancing the efficiency of data retrieval. The retrieved knowledge can be further reasoned upon to generate new hypotheses, which can then be interpreted to promote a deeper understanding of novel concepts. Together, these tasks made up the entire BKG system, driving healthcare advancements.

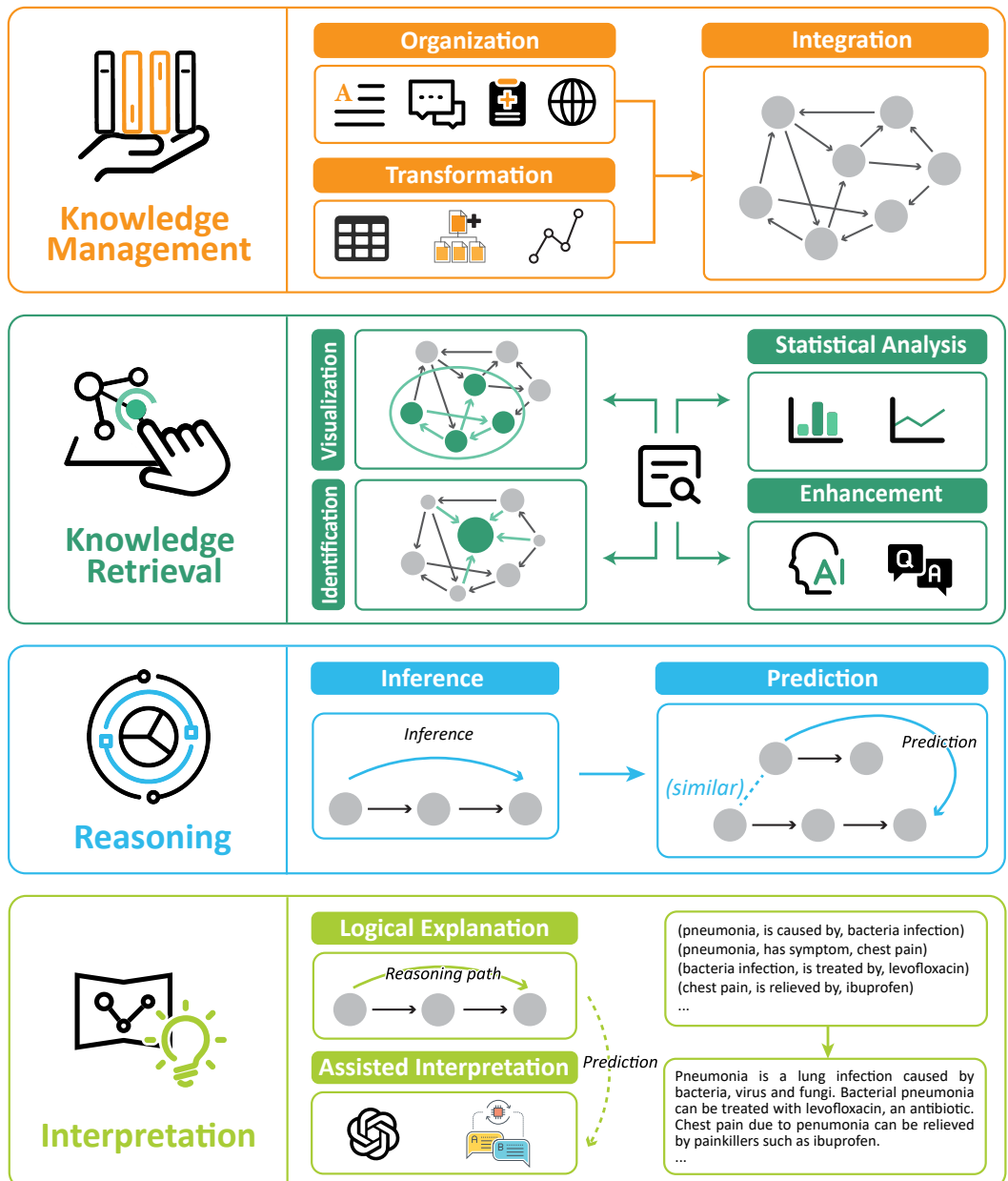

**Figure 3. BKG Tasks** Overview of BKG tasks, including knowledge management, retrieval, reasoning, and interpretation, which collectively enable diverse biomedical applications.

**Knowledge management**

As a specialized knowledge graph, BKG collects, organizes, and stores data with three subtasks: organization, transformation and integration. Through the organization of raw data and the transformation of existing datasets into standardized formats, resources become readily accessible for seamless integration.

**Organization**

Data organization refers to the process of structuring and classifying raw data, particularly unstructured data scattered across various resources. The amount of unstructured textual medical knowledge such as scientific literatures [89, 103] and patient registries [31, 33, 90] has massively increased due to information digitalization [104]. Knowledge extraction from these sources enrich and complete BKGs, supporting BKGs in organizing and combining texts from biomedical text sources [43, 45, 67, 84, 102, 105].

Sang et al. proposed SemKG [103] through mining biomedical literatures, where triples are extracted from PubMed and integrated in combination with the UMLS semantic network [106]. Similarly, Percha et al. developed a knowledge network by extracting chemical, gene and diseases from Medline abstracts through a distributional semantics-based approach [89]. EMR is another valuable medical source, but much essential data such as handwritten reports and doctor-patient dialogues, remain unstructured despite the structured workflow of EMR systems. To address this issue, Rotmensch et al. proposed a methodology for the automatic derivation of a disease-symptom KG based on EMR data, where disease and symptom entities were extracted from records such as chief complaint and clinical notes [90]. Gao et al. collected a batch of ophthalmic medical records including their matched medical images and reports to learn the association between biomedical entities [33]. Zheng et al. constructed a COVID-19 multi-modal BKG compiling text-converted doctor-patient dialogues and radiological images of patients, which is then further utilized for COVID-19 diagnosis [31].

**Transformation**

Data transformation is the process of converting heterogeneous structured data into a common standardized format [62, 107, 108]. The majority of biomedical structured data are in tabular form, mostly from public repositories such as GEO [109], DECIPHER [110] and MIMIC [111], but even tabular datasets obtained can be stored in different formats like CSV, JSON and XML. In order to manage the disparate datasets into a single database, transformation is required to standardize the dataset structures [42, 52, 54, 61, 72, 73, 82, 83, 90, 112, 113].

Nováček et al. performed the format conversion of a single biological dataset into a phosphorylation KG, where phosphorylation statements from the PhosphoSitePlus dataset were converted and compiled into the phosphorylation network [62]. Hasan et al. introduced a KG-based cancer registry management system for the Louisiana Tumor Registry dataset, linking tumor characteristics and treatment outcomes through networks [107]. Data transformation also includes format standardization between datasets from numerous sources. For instance, Cavalleri et al. constructed a KG representing interactions that involve RNA molecules, by mapping ontology to data sources [54]. Chandak et al. harmonized 20 structured datasets into a standardized format and defined a unique identifier for each node type, constructing PrimeKG [112]. Shang et al. transformed the EHR data into an RDF-type patient information graph, before integrating it with relevant clinical knowledge [82].

**Integration**

Data integration is the process of merging knowledge from multiple sources into a single BKG, essential for the modeling of complex biological system. Isolated datasets including raw data [1, 64, 107], tabular data [52, 82, 99, 107] and networks [42, 114, 115] can be connected, BKG can then be applied into healthcare researches for domain experts to explore biomedical issues with ease.

BKG integrates structured datasets from distinct databases. Himmelstein et al. compiled biomedical knowledge from 29 public data resources into Hetionet, encompassing information from millions of studies [99]; whereas Shang et al. merged real-world data and relevant clinical knowledge into an EHR-oriented KG [82]. Djeddi et al. built a BKG comprising PPIs and other biomedical entities, providing an illustration of connection between drugs and certain genes [114]; while Balbi et al. integrated a BKG with PPI graph and developed a domain knowledge-injected network [115]. Mohamed et al. generated a large scaled-BKG for DTI prediction by combining several individual KGs derived from several public datasets [42]. Delmas et al. introduced FORUM, a BKG built from life science databases and scientific literatures, demonstrating relations between molecules and biomedical concepts [64]. Santos et al. proposed the CKG which comprises clinical proteomics data from experimental data, public databases and literatures, encapsulating a massive amount of biomedical knowledge across publications [1].

**Knowledge retrieval**

As a retrieval platform, BKG enables direct information acquisition by accessing entities and relations. Four sub-tasks can be performed on the basis of its retrieval capability: data visualization, statistical analysis, knowledge identification and knowledge enhancement.

**Data visualization**

Data visualization refers to the demonstration of subgraphs or meta-paths of certain entities, illustrating the patterns of specific mechanisms within biological system. It provides a direct method to enhance the interpretation of the KG, and facilitates the process of knowledge retrieval by creating user-friendly interfaces for queries.

Knowledge regarding certain topics is displayed as subgraphs after retrieval to promote further analysis [1, 4, 5, 53, 66, 107, 116-124]. Jha et al. built GenomicsKG from integrated multi-omics data to visualize poly-omics data and facilitate the explanation of cancer mechanism [53]. Huang et al. proposed a BKG explainer module that transforms relevant subgraphs into drug-disease paths, along with an interactive tool displaying the selected drug's relative position as in the entire drug candidate subgraph [116]. Besides, Qian et al. visualized the information in their CVD Atlas KG through disease, trait and gene network [117]; Huang et al. visualized EMRs to assist diagnosis by enabling doctors to retrieve similar patient cases, reducing misdiagnosis risk [118].

**Statistical analysis**

Data retrieved from BKG can be analyzed using statistical methodologies to reveal patterns like disease stage distribution and clinical significance levels [1, 54, 91, 94, 97, 99, 107, 125-127]. It goes beyond common database functions as statistical analysis on graph structures provides additional information. These descriptive statistics are the result of BKG interpretation and summarization.

Santos et al. implemented an automated analysis pipeline along with their CKG database, where summarize characteristics and relevance of clinical proteomics entities [1]. Besides, Welivita et al. analyzed the distribution of affective states to examine the association between human emotions and distressing situations through a BKG [125]. Hasan et al. investigated the socio-economy factors of breast cancer in their registry KG, deriving age-specific incidence rates and its concentrated disadvantaged index distribution [107]. Furthermore, Chen et al. developed CTKG, where some of the nodes represent statistical methods utilized and their corresponding analysis outcomes [91]. These statistical analytics provide insights into the common connection between biomedical concepts.

**Knowledge identification**

Knowledge identification refers to the process of discovering existing yet unutilized opportunities, such as potential treatment approaches or possible disease pathogenesis. These opportunities often remain unexplored due to the lack of scientific findings transformation, where research achievements are published but not recognized by experts. Knowledge might not be apparent when viewed in isolation, thus BKG serves as a powerful tool to bridge this gap by demonstrating scientific findings clusters [1, 128] and common path lengths [107].

BKG compiled from literatures is an approach to prioritize these unexplored findings [67, 99, 103, 105], providing possible solutions such as alternative treatments and possible mechanism of an underexplained symptom. Santos et al. identified alternative treatment options for chemo refractory by prioritizing co-mentioned drug combinations, where drug-target-disease triples mentioned by numerous literatures are mined from the CKG [1]. Xu et al. proposed a PubMed KG to create node clusters such as potential drugs discovered by high-impact journals, and treatments identified by numerous articles [128]. Hence, aligned with the idea of scientific findings clustering, knowledge identification can also be interpreted as basic LBD [103, 129, 130]. Hasan et al. computed cancer treatment sequences that represent treatment process of an individual, by retrieving treatment paths from the cancer registry KG, enabling hierarchical structured-clinical queries [107]. BKG collects knowledge across various resources, extracting and organizing co-mentioned knowledge into single entities [131, 132], fully exploring associations between disparate biomedical literatures and real-life database.

**Knowledge enhancement**

On the basis of data retrieved from BKG, knowledge enhancement can be performed. BKG is a valuable background source for knowledge-aware applications, owing to its capability in integrating various domain-specific data. Thus, knowledge enhancement involves leveraging BKG as an external knowledge source for intelligent applications such as medical diagnosis [33, 58, 60, 82, 83, 133-135], and LLMs [20, 59, 136-140].

The compilation of medical theories and real-life data into a KG is an advantageous context source of medical diagnosis, as doctors use them as references for patients with similar symptoms. Gao et al. proposed an auxiliary diagnostic model on their multi-modal KG which compiles real-world ophthalmology medical data, automatically recommends similar previous cases for a new patient's condition [33]. Similarly, Liu et al. introduced a KG-based healthcare recommendation system for treatment suggestions [133], whereas Gong et al. proposed a medicine recommendation system through the utilization of a KG consisting of EMR end medical knowledge [83]. Zhang et al. constructed a BKG for automatic report generation, by capturing global visual information of the radiology images for disease classification [134].

BKG can also be the context source of LLMs. Training LLMs on BKG datasets can reduce the generation of non-factual answers from general unaided LLM, developing biomedical LLMs that can overcome medical language barriers, summarize clinical notes and even aid the evaluation of patient conditions [6, 49-51]. Soman et al. combined SPOKE BKG with LLM chat models to pretrain and encode knowledge into the biomedical LLM [136], while O'Neil et al. integrated the Monarch KG with LLM to enhance the accessibility of chat queries, improving the performance of healthcare intelligent question answering system [137]. Moreover, Jia et al. constructed a KG by crawling web pages, aiming to enhance patient's description regarding their condition and disease characteristics accurately [138]. Meanwhile, BKG improves biomedical entity representations by being the knowledge source. For example,

Zhang et al. incorporated OntoProtein with a BKG, integrating biological knowledge into protein representations [59]. Likewise, Kalifa et al. enhanced PLM by training it on the entire protein KG, enabling the model to capture broader relational nuances and dependencies, enriching protein representations [140].

**Knowledge reasoning**

Reasoning is the process of logically drawing conclusions or generating new knowledge based on existing knowledge. In BKG, reasoning comprises two distinct approaches: inference and prediction. Inference typically relies on complex statistical modeling, while prediction involves the learning process.

**Inference**

Inferencing can be defined as a complex solving process that identifies underlying possibilities in the graph. Rule-based inferencing [41, 73, 75, 89, 99] and probabilistic inferencing [141-143] are the two main methods of generating inferences [33, 58, 60, 67, 73, 86, 103, 107, 144], interpreting entity relationships to identify novel relation between indirectly linked entities. A common characteristic of these methods is that thresholds such as meta path length, extent of similarity measures and minimum probability of edge existence, can be predefined to support knowledge inferencing.

*Rule-based inference*

Rule-based inferences are conclusions made through a set of rules, including predefined paths [41, 73], entity similarity [75], and clustered patterns [89, 99]. Zhu et al. computed formulas to measure the drug-to-disease meta paths of their drug KG [73]; while Jiménez et al. generated inferences based on predefined meta paths to connect a drug to its corresponding disease [41]. Besides, similarity measures regarding entity information are considered as rule-based inferencing approach. Abdelaziz et al. compared drugs with a set of drug information derived-features such as chemical structure, side effects and drug target to predict similar drug pairs [75]. Furthermore, clustering of graph patterns is ubiquitous in rule-based inferencing. Percha et al. categorized groups of dependency paths based on their semantic similarity, which resulted in numerous clusters representing similar characteristics [89]; whereas Himmelstein et al. identified network patterns of Hetionet, distinguishing treatments groups from non-treatments groups [99].

*Probabilistic inference*

Hypotheses drawn from probability computations are considered as probabilistic inferences [141-143]. For instance, McCusker et al. developed the evidence-weighted BKG where relations are assigned with probability based on the methods used to create the assertion, supporting the identification of several drug candidates for melanoma through probabilistic analysis [142]. Ernst et al. narrowed the list of candidates using probabilistic weights derived from BKG [141], while Jiang et al. created a Markov network theory based-probabilistic model from weighted KG for medical diagnosis [143].

**Prediction**

Knowledge prediction, often referred to knowledge graph completion (KGC) aims to predict missing links and mine unknown facts [145, 146] to support hypotheses generation tasks such as drug candidate, diagnosis and risk prediction [31, 33, 44, 69, 96, 99, 105, 147-150]. Prediction techniques are broadly classified into rule-based prediction [41, 43, 93, 103, 113, 119, 151] and embedding-based prediction [36, 39, 42, 62, 73, 75, 83, 84, 91, 97, 116, 152-154].

*Rule-based prediction*

Rule-based predictions are conclusions inferred by directly utilizing data as input for statistical analysis and AI algorithms. Yamanishi et al. proposed several supervised statistical methods that reflects the drug-target interaction, namely nearest profile method, weighted profile method and bipartite graph learning method for drug-drug interaction (DDI) prediction [119]. Liu et al. proposed PoLo, a BKG multi-hop reasoning approach with logical rules in the form of meta paths [43]; while Jiménez et al. introduced a path-based drug repurposing method using random walks and reinforcement learning techniques and variational inference [41]. ML-based classification models are also highly utilized in the biomedical field. Tao et al. developed a health risk classification model by establishing a threshold for patient health boundaries, utilizing BKG [113]; likewise, Sang et al. trained a logistic regression model by learning semantic types of known drug therapies' paths [103].

*Embedding-based prediction*

KGE is a common approach to transform entities and relations of a KG into low-dimensional vectors [12, 37, 73], these embeddings can then be utilized to compute the similarity between entities, facilitating node and link predictions. Embedding techniques utilized for prediction can be generally classified into translational model [12, 13, 73, 75, 83, 91], semantic matching model [36, 42, 62, 74] and GNN model [13, 38, 39, 116, 152].

Translational models are structural embedding models that represent relationship as translations from head to tail entities [12, 13]. The similarity between entities and relations can be calculated by through embedding similarity. For instance, Chen et al. implemented TransE to generate nodes embedding of CTKG, cosine similarities between the condition and drug nodes are then calculated for the identification of 10 most similar pairs [91]. Abdelaziz et al. developed a large-scale similarity-based framework for DDI prediction using TransH, calculating cosine similarity between vectors to classify the most similar embeddings [75]. Gong et al. embedded different parts of the heterogeneous graph with TransR to learn the similarities between patient-medication relationships [83]. RESCAL [29], DistMult [28] and ComplEx [155] are three common semantic matching models. Nian et al. used ComplEx to produce corrupted triples as negative samples for training, improving the capability of KGC and identification of Alzheimer disease's potential treatment candidates [74]. Mohamed et al. developed TriModel, a tensor factorization model extending DistMult to learn efficient vector representations of drugs and targets, through iterative learning on training sets [42]. Furthermore, GNNs are KGE models that transform information of every node's neighborhood into a dense vector embedding [13, 38, 39, 152]. For instance, Huang et al. proposed TxGNN, a DL approach for drug predictions by incorporating GNN model for disease entity embedding and capture complex disease similarities [116]. Feng et al. obtained drug embeddings by GCN, which then become the input of the DNN model predictor for training to predict potential DDIs [152].

Apart from the three KGE methods, LLMs have emerged as an innovative link prediction approach that fully leverages the textual information and graph topology of KGs [48]. Xiao et al. proposed FuseLinker, a BKG link prediction framework leveraging both text and knowledge embeddings [154]. LLM is incorporated to generate text embeddings from BKG, the text embeddings are then fused with domain knowledge embeddings and ultimately become the input of GNN model that supports link prediction.

**Knowledge interpretation**

On the basis of new hypotheses or knowledge generated in the previous parts, knowledge interpretation provides an explanation by backtracking through the reasoning process, critical to support and validate the reliability of knowledge generated. Interpretation can be categorized based on how the predicted outcome is explained: logical explanation refers to an in-KG interpretation that displays the reasoning process within the graph's topology; whereas assisted interpretation is an out-of-KG interpretation that translates the reasoning process into texts with the help of artificial intelligence systems like LLM.

**Logical explanation**

Logical explanation illustrates the reasoning process by showing the original path crucial for prediction, represented as triple sets or subgraphs [38, 41, 72, 82, 97, 99, 116, 150, 151]. Explanatory paths significantly enhance the reliability of new knowledge as they display the actual associations between biological entities, allowing field experts to acquire a full understanding of the reasoning process.

Huang et al., developed an explainer module provides an overview of meta paths relevant to the prediction, displaying biological relationships crucial for its therapeutic predictions [116]. Shang et al. proposed an EHR-oriented KG providing graphical illustration of the reasoning, along with an explanation regarding the clinical significance of the prediction output [82]. Apart from that, Jiménez et al. proposed XG4Repo, highlighting its capability to provide prediction explanations, as well as generating Cypher queries for path acquisitions [41].

**Assisted interpretation**

Recently, LLMs are integrated with KG to aid in the explanation of a specific outcome [23, 137, 156-164]. The ability of LLMs for queries interpretation enhances the interpretability and reliability of a predicted outcome. It is a key element for healthcare intelligence, improving users' understanding of medical queries and helping doctors from different medical specialties to interpret a patient's condition.

Graph-to-text generation is an example of assisted interpretation, where top explanatory paths or sub-graphs crucial to the reasoned outcome are converted into text-based explanations. Lu et al. proposed a Clinical Retrieval-Augmented Generation (RAG) pipeline, where enriched medical knowledge can be converted into natural language to provide comprehensive answers with proper references [165-167]. Meanwhile, Pelletier et al. developed a RAG workflow RUGGED that provide answers and relevant literature evidences for queries [156]. O'Neil et al. introduced Phenomics Assistant, a chat-based interface for KG queries that integrates LLM with BKG, to answer queries with verified data and links to relevant sources [137].

## Applications

BKGs are developed to support a range of downstream applications in the biomedical field (**Figure 4**). From integrating diverse datasets across domains to performing multiple tasks on BKGs, various real-world applications can be effectively conducted. These applications often leverage one or more datasets and tasks within a single use case. Importantly, BKGs provide a robust foundation of prior knowledge, enhancing query platforms and empowering knowledge-aware applications.

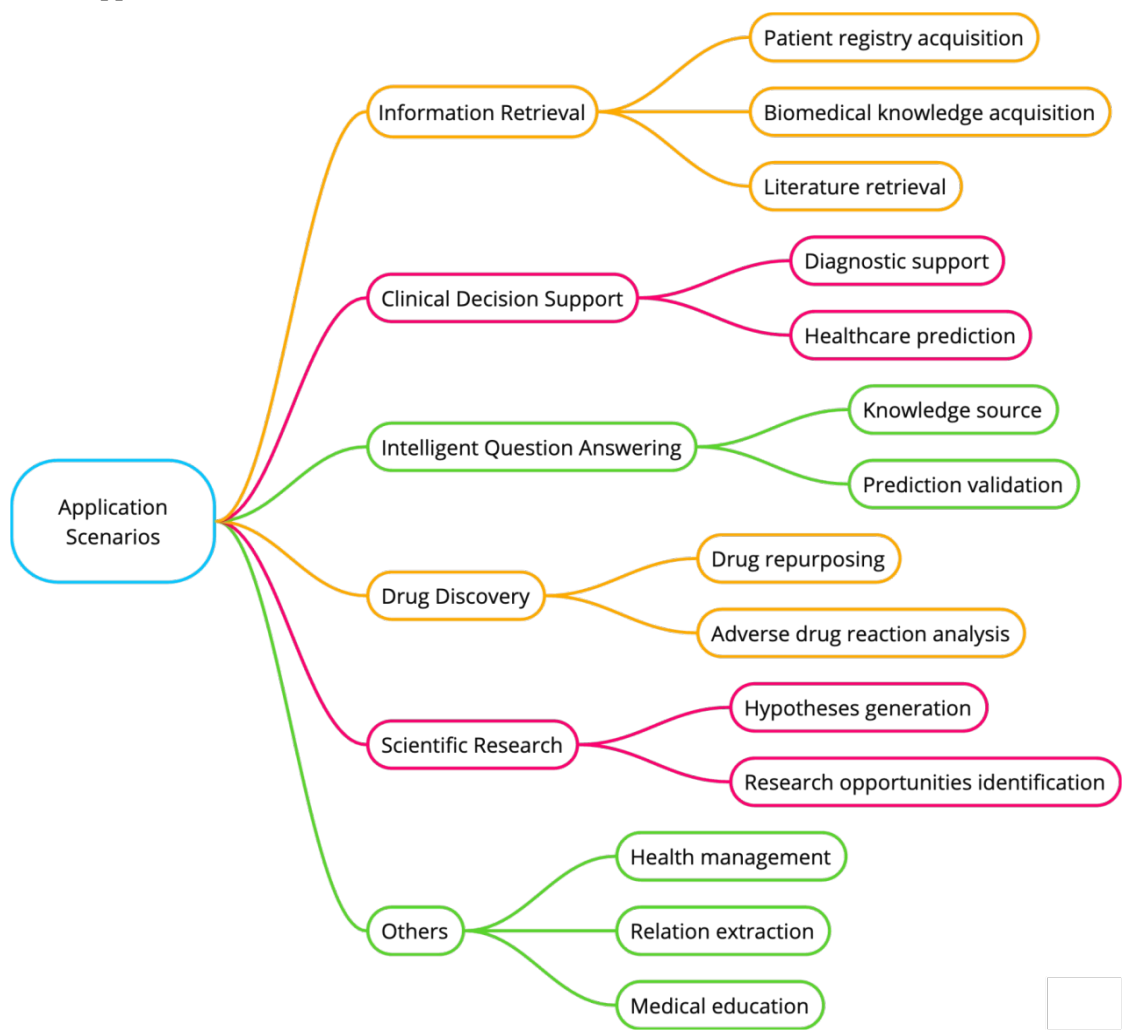

**Figure 4. BKG Applications.** An overview of the BKG utilization, highlighting the five main applications in the biomedical field.

## Information Retrieval

Knowledge retrieval is the basic function of BKGs serving as knowledge bases, where tasks such as visualization and statistical analysis provide additional context to enhance the retrieval process. Various biomedical retrieval systems have been developed to support user queries, including patient registry acquisition [82, 92, 107, 108, 168], biomedical knowledge acquisition [52, 84-86, 97, 99, 102, 112], and literature retrieval [128].

**Patient registry acquisition**

BKGs function as patient registry database by collecting and organizing patient-related data based on specific conditions and diagnoses. Hasan et al. constructed a BKG to manage cancer registry data, compiling demographic and treatment information of cancer patients [107]. Wang et al. proposed the personal health record (PHR) KG, which contains six aspects of patient data extracted from EMRs, improving individual healthcare management [168]. Yu et al. presented the integration of BKG into an AI chronic disease management system, enabling the retrieval of specific cases as diagnosis references [108].

**Biomedical knowledge acquisition**

A significant number of BKGs incorporate biomedical concepts and theories, such as cellular mechanisms and pathological pathways. Hetionet [99], PharmKG [97] and PrimeKG [112] BKGs that provide cross-domain biomedical knowledge. Focusing on human genome, Feng et al. [52] developed GenomicKB to enable retrieval of information related to genome, epigenome, transcriptome and 4D nucleome. Furthermore, disease-specific KGs, such as SDKG-11 [84] and KGHC [85], facilitate the acquisition of knowledge related to specific diseases.

**Literature retrieval**

Xu et al. developed PubMed KG (PKG) that collects abstracts and author-related information [128]. PubMed biomedical literatures, author names, affiliation history and other researcher-centric activities are incorporated in the BKG to illustrate relationships between biomedical entities and their current research trend. PKG enables healthcare experts to efficiently retrieve biomedical studies for research purposes, supporting decision-making by considering the scientific contributions and academic success of researchers.

**Clinical Decision Support**

Clinical decision support (CDS) is a key application scenario of BKGs [169]. By integrating medical knowledge with patient-specific information, BKGs improve healthcare diagnosis and assist doctors in making decisions, enhancing the field of precision medicine. Clinical decision support system CDSS relies on BKGs as both a knowledge source and a prediction platform, enabling diagnostic support [134, 135] and healthcare prediction [31, 33, 83, 86, 87, 92, 93, 102, 113, 133, 148, 150, 170].

**Diagnostic support**

Medical knowledge stored in BKGs enables CDSS to assist in detecting biosignals and symptoms, supporting doctors in their diagnosis procedure [171, 172]. Sekuboyina et al. performed chest X-ray classification with their radiological KG by conducting link predictions within the BKG [135]. Likewise, Zhang et al. presented a radiology report generation framework by incorporating a BKG related to chest findings, enhancing the interpretation of medical images [134].

**Healthcare prediction**

Ma et al. proposed KAME, a model which learns embeddings from BKG to predict future health information of patients based on their historic data [87]. Tao et al. constructed a health BKG to develop a health risk classification model, predicting associations between a patient's symptoms and diseases [113]. Moreover, BKG-integrated CDSS provides medication recommendations to reduce medication errors [169]. Bhoi et al. developed PREMIER, a personalized medication recommendation system that models drug interactions and drug co-occurrences of individuals [150]. Shang et al. presented GAMENet, a model incorporating a DDI KG, to recommend medication combinations for patients with complex health issues [148].

**Intelligent Question Answering**

The advancements of NLP techniques and LLMs have driven the rise of intelligent question-answering systems, these systems process questions and generate answers based on semantic similarity [6]. The role of BKGs in intelligent question answering lies in providing contextual information [86, 125, 136-139, 173-177] and validating LLM predictions [48, 178-180], thereby preventing hallucinations in biomedical chat models.

**Knowledge Source**

BKGs play a crucial role in supporting intelligent question-answering systems by assisting answer generations [86, 125, 136-139, 173-177, 181], understanding user's queries [137] and aiding patient condition description [138]. Yasunaga et al. developed DRAGON to pretrain biomedical language models using BKG, enabling question answering across the biomedical field [139]. BKG is also leveraged to interpret queries in natural language, helping non-specialists retrieve relevant information, as demonstrated by O'Neil et al. [137]. Furthermore, Jia et al. proposed DKDR, a disease diagnosis model that integrates BKG to generate accurate descriptions of patient conditions for doctors [138].

**Prediction validation**

A common challenge with AI chatbots is the generation of inaccurate or even incorrect answers. BKGs, as repositories of factual knowledge, can be used to probe the knowledge contained in existing language models (LMs). Sung et al. developed the BIOLAMA benchmark, which consists of biomedical factual knowledge triples for LM probing [178]. LM predicts missing entities, and the resulting knowledge can be validated against a medical KG to ensure the accuracy and reliability of knowledge stored within the LM.

**Drug Discovery**

BKGs play a vital role in drug discovery by performing link predictions based on accumulated knowledge to identify potential drug and target candidates. The application of BKGs in drug discovery encompasses two main areas: drug repurposing [38, 42, 44, 67-69, 73, 74, 84, 97-99, 105, 116, 142, 182] and adverse drug reaction (ADR) analysis [39, 75, 76, 95, 183].

**Drug repurposing**

DTI and DDA are two common types of interactions modeled by BKGs, which play a key role in drug repurposing [4, 9]. Huang et al. proposed TxGNN, a model for zero-shot drug repurposing trained on a medical KG, to rank potential drug indications and contraindications for various diseases [116]. Focusing on COVID-19, Zhang et al. predicted drug candidates through KGC techniques [67]. By combining BKGs with recommendation system, DTI can also be predicted to facilitate drug repurposing efforts, as illustrated by Ye et al. [182].

**ADR analysis**

In BKGs, DDIs as the main cause of ADRs, form the foundation of ADR analysis [4, 9]. Bean et al. developed a ML algorithm on the basis of their BKG, to classify the causes of ADR and assist in detecting novel ADRs [76]. Similarly, Wang et al. identified potential ADRs of antitumor drugs using their Tumor-Biomarker KG, demonstrating the capability of BKGs in predicting ADR occurrences and exploring their mechanisms [183].

**Scientific Research**

The knowledge foundation of BKGs highlights their advantages in providing accurate information across various domains. Hypotheses generation [1, 33, 60, 64, 85, 94, 96, 101, 102, 151, 170, 179, 184, 185] and research opportunities identification [1, 107] are two primary areas where BKGs are utilized in biomedical research.

**Hypotheses generation**

BKGs generate high-potential hypotheses for biomedical researches based on the theoretical knowledge embedded within the graph. Karim et al. [179] and Feng et al. [184] developed BKGs for cancer-specific biomarker discovery. Predictions made by these BKGs reveal possible associations between cancer and biological molecules, providing critical diagnostic information. Similarly, Pelletier et al. [151] and Renaux et al. [94] leveraged BKGs for pathology exploration, investigating organellar pathways and gene interactions to uncover the causes and mechanisms of diseases. The biomarkers and pathological insights predicted serve as generated hypotheses that require wet-lab experiments validation.

**Research opportunities identification**

BKG consolidate information across biomedical domains, including scientific findings published in biomedical literatures. These research achievements often remain underutilized, BKGs therefore bridge the gap between research and practical applications, by highlighting existing opportunities. The underlying possibilities can be future research directions, which can be transformed into real-world applications upon validation. To illustrate, Santos et al. introduced a clinical KG (CKG) where alternative treatments are prioritized by displaying co-mentions of relevant publications [1]. Furthermore, the cancer registry KG proposed by Hasan et al. highlighted racial disparities among TNBC women, providing a promising research area for further exploration of the disease [107].

**Others**

BKG applications extend beyond the five areas mentioned above and can be leveraged for many other knowledge-aware applications, such as health management [144, 186], relation extraction (RE) [16, 17, 20, 187-189], and medical education [190, 191].

Disease prevention is a crucial aspect of healthcare. Huang et al. proposed a healthy diet KG for healthcare management, compiling information about food, nutrient elements, and symptoms to address the dietary requirements of different people [144]. Dang et al. developed GENA, a BKG that models nutrition-mental health associations by extracting abstracts related to chemicals, food, and health [186]. Apart from that, BKGs are leveraged as text-mining tools to perform biomedical RE tasks due to their semantic-based network structure, supporting LBD. Sousa et al. integrated BKGs into biomedical RE system, identifying and classifying complex relationships among biomedical entities [20]. Roy et al. developed three ClinicalBERT models that integrate the UMLS KG into pre-trained language models [17]; while Sun et al. designed KGBReF to discover emotion-probiotics relation through BKG [187]. Another noteworthy application of BKGs is in medical education systems. Ettorre et al. introduced SIDES, a web-based platform that provides personalized learning services by extracting supplementary knowledge from the OntoSIDES BKG [190]. Likewise, Ansong et al. improved the construction of disease diagnosis features in current medical training systems through BKGs, empowering the advancement of educational technology [191].

In summary, BKG applications extend beyond the scenarios listed above, offering versatile solutions in the biomedical field.

# Resources

BKGs offer valuable support for researchers across various biomedical tasks. This section provides an overview of biomedical databases, existing BKGs, and tools specifically designed for KG construction.

**Biomedical Databases**

Biomedical databases serve as foundational knowledge sources for BKG datasets, organizing biomedical information across the omics, pharmacology, and medical domains. These heterogeneous datasets represent the biological system and include components such as genes, transcripts, proteins, chemicals, drugs, diseases, and side effects. Meanwhile, biomedical ontology bases are also considered databases but they differ from traditional databases by standardizing representation, definitions, and naming conventions for biomedical entities, ensuring consistency in scientific concepts. **Table 1** provides a list of common biomedical databases along with details on their data types.

**Existing BKGs**

The development of BKGs for research and various applications has grown significantly in recent years, driven by their capabilities as powerful semantic networks. These BKGs address distinct aspects of the biomedical field, incorporating diverse types of data. Existing BKGs vary in scale, data sources, and domain focus, depending on their intended applications. Furthermore, KGE is a crucial technique to capture and represent relational information in BKGs, but its presence and implementation vary across different BKG systems. **Table 2** presents a list of existing BKGs that can be directly utilized for research and practical applications.

**KG Tools**

The versatility of BKGs, ranging from static to dynamic, unweighted to weighted, and general to specific, demonstrates their highly adaptable nature. The structure of a BKG can be tailored to meet specific research objectives and practical application requirements. With the ongoing progress in technological development, researchers can now construct customized KGs optimized for their work. To support this, we provide a list of KG tools specifically designed for automated KG construction and KG analysis in **Table 3.**

**Table 1. Biomedical Databases. A comparative overview of dataset coverage across common biomedical databases.**

| | Database | Dataset coverage | | | | | | | |
|---|---|---|---|---|---|---|---|---|---|
| | | Gene | Transcript | Protein | Chemical | Drug | Disease | Pathway | Side effect |
| Omics | KEGG | ✓ | | ✓ | | ✓ | ✓ | ✓ | |
| | PharmGKB | ✓ | | ✓ | ✓ | ✓ | ✓ | | |
| | Ensembl | ✓ | ✓ | | | | | | |
| | Entrez Gene | ✓ | | | | | | | |
| | OMIM | ✓ | | ✓ | | | ✓ | | |
| | RNAcentral | | ✓ | | | | | | |
| | UniProtKB | ✓ | | ✓ | | ✓ | ✓ | ✓ | |
| | TTD | ✓ | | ✓ | ✓ | ✓ | | | |
| | HPA | ✓ | | ✓ | | | | | |
| | InterPro | | | ✓ | | | | | |
| | STRING | ✓ | | ✓ | | | | | |
| | Reactome | ✓ | | ✓ | | | ✓ | ✓ | |
| | WikiPathways | | | | | | ✓ | ✓ | |
| | *GO | ✓ | | ✓ | | | ✓ | ✓ | |
| | *PRO | ✓ | | ✓ | | | ✓ | ✓ | |
| Pharma-cology | PubChem | | | | ✓ | | | | |
| | ChEMBL | | | ✓ | ✓ | ✓ | ✓ | | |
| | DrugBank | ✓ | | | ✓ | ✓ | | ✓ | |
| | DrugCentral | | | | | ✓ | ✓ | | |
| | CTD | ✓ | | ✓ | ✓ | ✓ | ✓ | ✓ | |
| | BindingDB | | | ✓ | ✓ | ✓ | | | |
| | Supertarget | | | ✓ | | ✓ | | | |
| | SIDER | | | | ✓ | ✓ | | | ✓ |
| | TWOSIDES | | | | | ✓ | | | ✓ |
| | *DRON | | | | | ✓ | | ✓ | |
| Medical | DISEASES | ✓ | | | | | ✓ | | |
| | DECIPHER | ✓ | | | | | ✓ | | |
| | CVD Atlas | | | | | | ✓ | | |
| | MIMIC-IV | | | | | | ✓ | | |
| | SemMedDB | ✓ | | | | | ✓ | | |
| | *UMLS | | | | | | ✓ | | |
| | *DO | | | | | | ✓ | | |
| | *MeSH | ✓ | | | | | ✓ | | |
| | *SNOMED CT | | | | | | ✓ | | |
| | *MONDO | | | | | | ✓ | | |
| | *HPO | | | | | | ✓ | | |
| | *ORPHANET | | | | | ✓ | ✓ | | |

*represents ontology knowledge bases*

**Table 2. Biomedical knowledge graphs.** An analysis of BKGs across the domains, highlighting their data sources, dataset statistics, utilization of KG embeddings, and main applications.

| | BKG | Data Sources | Dataset statistics | | | | KGE |
|---|---|---|---|---|---|---|---|
| | | | Entity Types | Relation Types | Entities | Relations | |
| Multi-Domain | PharmKG [97] | OMIM, PharmGKB, … (6) | 3 | 29 | 7,601 | 500,958 | ✓ |
| | OpenBioLink [192] | KEGG, DrugCentral, … (13) | 7 | 30 | - | - | ✓ |
| | BioKG [100] | UniProtKB, REACTOME, … (14) | 5 | 10 | - | - | ✗ |
| | ROBOKOP KG [193] | DrugBank, HPO, … (26) | 54 | 1064 | ~8.6M | ~130.4M | ✗ |
| | KnowLife [141] | MeSH, PubMed, … | - | 14 | 214,000 | 78,000 | ✗ |
| | BIOKGLM [194] | UniProtKB, SemMedDB, … | 5 | 11 | ~502.1k | ~96.5M | ✓ |
| | PrimeKG [112] | CTD, SIDER, … (20) | 10 | 30 | 129,375 | 4,050,249 | ✓ |
| | Hetionet [99] | ChEMBL, BindingDB, … (29) | 11 | 24 | 47,031 | 2,250,197 | ✗ |
| | GNBR [105] | Orphanet, DrugCentral, … | 3 | 32 | 63,252 | 583,685 | ✓ |
| | DRKG [195] | DrugBank, IntAct, … (6) | 13 | 107 | 97,238 | 5,874,261 | ✓ |
| | MegaKG [72] | GO, TTD, … (23) | 7 | 20 | 188, 844 | 9,165,855 | ✓ |
| | BIKG [68] | ChEMBL, GO, ... | 22 | 59 | ~11M | ~118M | ✓ |
| | TarKG [98] | MESH, Hetionet, … | 15 | 171 | 1,143,313 | 32,806,467 | ✓ |
| Omics | GenomicsKG [53] | TCGA, COSMIC, ... | - | - | - | - | ✗ |
| | GenomicKB [52] | GENCODE, GO, … | 26 | 13 | - | - | ✗ |
| | BOCK [94] | OLIDA, Mentha, … | 10 | 17 | 158,964 | 2,659,064 | ✓ |
| | RNA-KG [54] | HPO, MONDO, … | 66 | 355 | 673,825 | 12,692,212 | ✓ |
| | ProteinKG65 [63] | UniProt, GO, … | - | - | 614,099 | - | ✗ |
| | IDP-KG [61] | ELIXIR IDP Community, DisProt, … | - | - | - | - | ✗ |
| Pharma-cology | DrugMechDB [66] | DrugBank, UniProt, … | 14 | 71 | 32,588 | 32,349 | ✗ |
| | OREGANO [70] | PharmGKB, SIDER, … (8) | 11 | 19 | 88,937 | 824,231 | ✓ |
| | GP-KG [69] | MGI, HPO, … (8) | 7 | 9 | 61,146 | 1,246,726 | ✓ |
| | IPM-KG [65] | DrugBank, CTD, … (4) | 10 | 17 | 151,631 | 420,126 | ✓ |
| | WATRIMed [196] | WATRIMed Plant database WATRIMed Plant Ontology | 13 | 36 | | | ✗ |
| | TBKG [183] | MEDLINE, UMLS, … | 4 | 6 | 6,291 | 139,254 | ✗ |
| Medical | CKG [1] | UniProt, DISEASES, … (25) | 36 | 47 | ~20M | ~220M | ✗ |
| | CTKG [91] | AACT | 18 | 21 | 1,496,684 | 3,667,750 | ✓ |
| | Rare diseases KG [101] | Inxight Drugs, GARD… (34) | 10 | 17 | 3,819,623 | 84,223,681 | ✗ |
| | SDKG-11 [84] | MirBase, ChEMBI, … (7) | 7 | 67 | 165,062 | - | ✓ |
| | KGHC [85] | SemMedDB, PubMed, … | 9 | 22 | 5,028 | 13,296 | ✓ |
| | KDKG [102] | DrugBank, SIDER, … (7) | - | - | - | 10,146,311 | ✗ |

**Table 3. BKG Tools. A compilation of tools for KG construction and analysis, along with their descriptions.**

| Name | Type | Description |
|---|---|---|
| Neo4J | Graph database | A system for graph database management storing nodes, edges, and their attributes for knowledge graphs. |
| ArangoDB | Graph database | A graph database system to store multi-modal datasets, encompassing functions such as analysis and visualization for knowledge graphs. |
| PheKnowLator | Construction | A semantic ecosystem with KG construction resources (data preparation APIs, analysis tools, KG benchmarks). |
| Neo4j LLM Knowledge Graph Builder | Construction | An online application for text-to-graph conversion of unstructured data, achieved without coding. |
| KGTK | Construction | A library of data-science centric tools and methods for various KG operations (construction, transformation, analyzation, validation, filtering etc.). |
| SemaRep | Extraction | An NLP system for semantic relations extraction from biomedical text using linguistic principles and UMLS knowledge. |
| TNNT | Extraction | A NLP-NER toolkit for the automatic extraction of categorized name entities from unstructured data. |
| DeepKE | Extraction | A deep-learning based toolkit for entity, relation and attribute extraction. |
| KGPrune | Subgraph extraction | A web browser and API to extract Wikidata subgraph related to interested entities. |
| PBG | Embeddings | A system to learn graph embeddings for large graphs with more than 100,000 nodes. |
| graph-tool | Graph analytics | A Python module for the manipulation and statistical analysis of graphs, supporting visualization and various network functions. |
| NetworkX | Graph analytics | A Python package for graph implementation, network analysis, network algorithms computation etc. |

# Limitations

Though BKGs provide powerful tools for representing and analyzing complex biological systems, several limitations still impact their effectiveness and widespread adoption. These challenges span technical, semantic, and practical dimensions, ranging from fundamental data integration issues to the complexities of maintaining up-to-date knowledge in a rapidly evolving field. Understanding these limitations is crucial for both users and developers of BKGs, as it helps inform future research directions and improvements in the field.

## Data Integration

The integration of biomedical data presents significant hurdles due to its inherent complexity and diversity. Data heterogeneity manifests across multiple dimensions: structural (different data formats and schemas), semantic (varying terminologies and definitions), and qualitative (inconsistent data quality standards). Source data often suffers from incompleteness, inconsistencies, and varying levels of reliability. Moreover, the lack of standardized data collection protocols across different institutions and laboratories further complicates the integration process, potentially compromising the overall quality and utility of the BKG. The challenge of harmonizing data from diverse sources while maintaining data integrity and provenance remains a significant obstacle in the development of comprehensive BKGs.

## Semantic Ambiguity and Standardization

Despite extensive efforts to standardize biomedical terminology through initiatives like UMLS, GO, and SNOMED CT, semantic ambiguity remains a persistent challenge. The biomedical domain is particularly susceptible to terminology variations due to its complex nomenclature system and the evolution of scientific understanding. Multiple naming conventions often exist for the same biological entities, and terms can have context-dependent interpretations. The abundant use of synonyms and acronyms, coupled with inconsistent usage of ontologies across different research communities, creates significant challenges in maintaining semantic consistency. These semantic challenges can lead to misinterpretation of data and complicate cross-domain knowledge integration, ultimately affecting the reliability and utility of the BKG.

## Scalability

As biomedical knowledge continues to grow exponentially, scalability has become a critical concern in BKG implementation and maintenance. The computational complexity of graph queries across large-scale networks poses significant challenges, particularly when real-time responses are required. Storage requirements for maintaining historical versions and provenance information continue to grow, creating resource constraints. Performance bottlenecks in query processing and the resource-intensive nature of graph analytics and inference operations often necessitate trade-offs between comprehensiveness and performance. These scalability issues can significantly impact the practical utility of large-scale BKGs, particularly in resource-constrained environments.

### Interpretability

While BKGs excel at representing complex relationships, their interpretability remains a significant challenge. Users often struggle with understanding the reasoning behind derived relationships and navigating complex network structures. The interpretation of probabilistic or uncertain relationships presents additional complexity, particularly when dealing with conflicting evidence or incomplete information. The challenge of efficiently accessing relevant information and comprehending large-scale network patterns is compounded by the lack of intuitive visualization and exploration tools. These interpretability issues can limit the practical utility of BKGs for end-users, particularly those without extensive technical expertise.

## Trends and Opportunities

### Current Trends

Modern BKGs are increasingly incorporating multi-omics data to capture complex relationships within biological systems. This multi-dimensional integration enables a more nuanced perspective on disease mechanisms and biological processes, offering a comprehensive overview of cellular and molecular interactions. Concurrently, the shift toward multimodal representations has led to the inclusion of diverse data types such as medical imaging, clinical notes, molecular structures, and longitudinal patient data. By encompassing these varied sources, BKGs provide richer contextual insights and support more robust analyses of biomedical knowledge.

Recent advances in large language models have further opened new opportunities for the construction and use of BKGs. These models are now being employed for automated knowledge extraction from scientific literature, more precise entity recognition, and more accurate relationship mapping, leading to sophisticated systems for biomedical knowledge representation and reasoning. Taken together, the incorporation of multi-omics data, multimodal information, and large language model capabilities positions BKGs as powerful tools for driving innovation in both biomedical research and clinical practice.

### Emerging Opportunities

BKGs present unprecedented opportunities for accelerating research by enabling researchers to identify novel connections among biological entities, predict potential drug targets, and gain a more comprehensive understanding of disease mechanisms. This capacity to rapidly synthesize and analyze vast biomedical datasets significantly reduces the time and resources required for scientific discovery. In drug discovery and development, BKGs are revolutionizing the process through more efficient identification of drug candidates, a deeper appreciation of drug-target interactions, and improved recognition of potential side effects. Their ability to integrate diverse data types and leverage AI-powered analyses also supports drug repurposing efforts by revealing new applications for existing pharmaceuticals. Furthermore, BKGs hold transformative potential in clinical settings, where they combine patient-specific data with extensive biomedical knowledge to deliver context-aware decision support. By guiding diagnosis, treatment planning, and outcome prediction through both retrospective information and real-time patient data, these systems enhance precision medicine strategies and enable clinicians to tailor interventions based on individual molecular profiles and clinical characteristics.

## Conclusion

BKGs have become essential tools for integrating and analyzing diverse biomedical data, enabling advancements in drug discovery, disease understanding, and clinical decision support. This survey highlights their transformative potential across domains such as multi-omics, pharmacology, and healthcare, while systematically summarizing current tasks and applications. Despite significant progress, challenges persist in data integration, scalability, and interpretability. Emerging approaches, including graph neural networks and the integration of large language models, present promising solutions to enhance reasoning and discovery capabilities. Future research should address these challenges by leveraging multimodal data and fostering collaborative innovations to fully realize the potential of BKGs in advancing precision medicine and translational research.

## Author Contributions

Conceptualization: J.W., Y.L.. Analysis: J.W., Y.L., S.G.. Project administration: J.W. Writing: J.W., Y.L., S.G., X.Z.. All authors discussed the results and reviewed the manuscript.

## Competing Interests statement

The authors declare no competing financial interests.

## Data availability

The paper is supplemented with a Github repository for the surveyed works and resources: https://github.com/YuxingLu613/awesome-biomedical-knowledge-graphs.